\documentclass[final,5p,times,nopreprintline,
]{elsarticle} 
\journal{Neural Networks}

\usepackage[english]{babel}
\usepackage[T1]{fontenc}
\usepackage[utf8]{inputenc}
\usepackage{lmodern}
\usepackage{microtype}
\usepackage{balance}

\usepackage{amsmath,amssymb,bm}
\usepackage{braket}

\usepackage{graphicx}
\usepackage[dvipsnames]{xcolor}
\usepackage{dcolumn}

\usepackage{natbib}
\usepackage[hidelinks]{hyperref}

\newcommand{\EqRef}[1]{Eq.~\eqref{#1}}
\newcommand{\FigRef}[1]{Fig.~\ref{#1}}

\begin{document}


\begin{frontmatter}

\title{Cross-Scale Reservoir Computing for large spatio-temporal forecasting and modeling}

\author[cref,iss,torvergata]{Nicola Alboré\corref{cor1}}
\ead{nicola.albore@cref.it}

\author[cref,iss,santalucia]{Gabriele Di Antonio\corref{cor1}}
\ead{gabriele.diantonio@cref.it}

\author[cref]{Fabrizio Coccetti}

\author[cref,romatre]{Andrea Gabrielli}

\cortext[cor1]{These authors contributed equally to this work.}

\address[cref]{Research Center ``Enrico Fermi'', Rome, Italy}
\address[iss]{National Center for Radiation Protection and Computational Physics, Italian National Institute of Health, Rome, Italy}
\address[torvergata]{Department of Physics, ``Tor Vergata'' University, Rome, Italy}
\address[santalucia]{IRCCS Santa Lucia Foundation, Rome, Italy}
\address[romatre]{Department of Civil, Computer Science and Aeronautical Technologies Engineering, ``Roma Tre'' University, Rome, Italy}

\begin{abstract}
We propose a new reservoir computing method for forecasting high-resolution spatiotemporal datasets. By combining multi-resolution inputs from coarser to finer layers, our architecture better captures both local and global dynamics. Applied to Sea Surface Temperature data, it outperforms standard parallel reservoir models in long-term forecasting, demonstrating the effectiveness of cross-layers coupling in improving predictive accuracy. Finally, we show that the optimal network dynamics in each layer become increasingly linear, revealing the slow modes propagated to subsequent layers.
\end{abstract}


\end{frontmatter}


\noindent
Complex spatio-temporal systems govern phenomena across multiple scientific domains, from atmospheric dynamics to plasma physics. These systems universally exhibit behavior across diverse timescales, presenting fundamental challenges for both traditional equation-based modeling and contemporary machine learning approaches. Although conventional approaches rely on detailed knowledge of the governing equations \cite{duraisamy2019turbulence, kalnay2003atmospheric, bauer2015quiet}, recent advances in data-driven modeling offer promising alternatives for systems where such knowledge remains incomplete or computationally intractable \cite{raissi2019physics, pathak2018model, price2025probabilistic}.

Reservoir computing (RC) has emerged as a powerful paradigm for modeling nonlinear dynamical systems, demonstrating remarkable efficacy in forecasting chaotic behavior with minimal computational overhead \cite{Jaeger2004, lukovsevivcius2009reservoir}. Standard implementations, however, typically struggle to simultaneously capture dynamics across disparate timescales, particularly when predicting systems with significant separation between fast and slow processes.
In recent years, new reservoir architectures have been proposed to address the limitations of the classical formulation \cite{gallicchio2017deep,pathak2018model,gauthier2021next, manneschi2021exploiting, kim2023neural, yan2024emerging}.
In this Letter, we turn our attention to the analysis of large-scale spatiotemporal datasets, where the approach of \cite{pathak2018model} remains the key reference. In that work, and in related studies, the domain is partitioned into grids of locally interacting reservoirs \cite{pathak2018model, arcomano2020atmospheric}. However, these designs do not account for the long-range dependencies essential for long-term forecasts. Inspired by physics-based treatments of long-range interactions in large N-body systems \cite{barnes1986hierarchical}, our architecture uses lower resolution layers to inform higher resolution reservoirs about long distances (\FigRef{fig:1}).

Application on Sea Surface Temperature (SST) dynamics demonstrate improved long-term reproduction compared to single-layer models (\FigRef{fig:2}).
This result comes mainly from the cross-scale underlying structure. Low-resolution levels filter out fast modes and emphasizing slow, coherent dynamics, helping the higher detailed layers to better dissect such information.

Finally, we demonstrate that the optimal RNN dynamics in each layer become increasingly linear as the resolution of the corresponding layer decreases. This results in an interpretable decomposition of the dynamical system into linearly evolving modes, revealing the slow modes propagated to subsequent layers.

\section{A multi-scale reservoir computing structure} 
\subsection{Reservoir Computing Framework}
We begin by describing the basic formulation of reservoir computing as used in this study. The framework consists of a high-dimensional, nonlinear dynamical system, namely the reservoir $\mathcal{R}$ which is driven by a lower-dimensional input signal. Let $\mathbf{u}(t) \in \mathbb{R}^{D_{\text{in}}}$ denote the input vector at time $t$, and let $\mathbf{r}(t) \in \mathbb{R}^{D_r}$ be the internal state of the reservoir.
The evolution of the reservoir state is governed by an ODE, in which the force field depends nonlinearly on the current state $\mathbf{r}(t)$ and the input $\mathbf{u}(t)$ via
\begin{equation}
    \frac{d\mathbf{r}(t)}{dt} = \mathcal{G} \left[ \mathbf{W}\mathbf{r}(t), \mathbf{W}_{\text{in}} \mathbf{u}(t) \right],
    \label{eq:reservoir-dynamics}
\end{equation}
where $\mathcal{G}$ is a nonlinear gain function, typically acting component-wise, and $\mathbf{W}_{\text{in}} \in \mathbb{R}^{D_r \times D_{\text{in}}}$ is a fixed matrix that projects the input into the reservoir state space. The matrix $\mathbf{W}_{\text{in}}$ is usually chosen randomly and kept fixed throughout training and prediction. In the specific implementation considered here, $\mathcal{G}=\tanh(\cdot) / \tau$ and $\mathbf{W} \in \mathbb{R}^{D_r \times D_r}$ is a matrix whose elements are sampled from a gaussian distribution with 0 mean and standard deviation $g/\sqrt{p D_r}$ with $p$ being the sparsity. 


If the reservoir dynamics settle into a subspace where responder states $\mathbf r(t)$ are constrained by the driver $\mathbf u(t)$, a synchronization map links the two. Extra dimensions in $r(t)$ may encode redundant or delayed information, preserving an echo of past information \cite{Jaeger2004}. This embedding is used to generate an output $\mathbf{v}(t) \in \mathbb{R}^{D_{\text{out}}}$ via a linear readout
\begin{equation}
    \mathbf{v}(t) = \mathbf{W}_{\text{out}} \, \boldsymbol{\psi}(\mathbf{r}(t)),
    \label{eq:readout}
\end{equation}
where $\mathbf{W}_{\text{out}} \in \mathbb{R}^{D_{\text{out}} \times D_{\psi}}$ is the matrix of trainable output weights and $\boldsymbol{\psi}(\mathbf{r})$ is a fixed set of features, often simply linear $\boldsymbol{\psi}(\mathbf{r}) = \mathbf{r}$, but quadratic or higher-order extensions are commonly used.
Training records reservoir states $\mathbf{r}(t)$ for $t \in [-T,0]$ under inputs $\mathbf{u}(t)$ and targets $\mathbf{v}_d(t)$. The weights $\mathbf{W}_{\text{out}}$ are then chosen to minimize the mean-squared error between $\mathbf{v}(t)$ and $\mathbf{v}_d(t)$ across the training interval, typically using ridge regression.
After training, the reservoir can be used for prediction. If the target is the driver itself, the dynamics describes an autonomous system. The input $\mathbf{u}(t)$ in Eq.~\eqref{eq:reservoir-dynamics} is replaced by the output $\mathbf{v}(t)$ of Eq.~\eqref{eq:readout}, giving
\begin{equation}
    \frac{d\mathbf{r}(t)}{dt} = \mathcal{G} \left[ \mathbf{W}\mathbf{r}(t), \mathbf{W}_{\text{in}} \mathbf{W}_{\text{out}} \, \boldsymbol{\psi}(\mathbf{r}(t) \right],
    \label{eq:reservoir-dynamics-2}
\end{equation}
which then generates predictions at each timestep $t > 0$ via the mapping in Eq.~\eqref{eq:readout}.

\subsection{Parallel Reservoir Computing}

When applying reservoir computing to systems characterized by large spatial domain, directly feeding the full input vector $\mathbf{u}(t) \in \mathbb{R}^{D_{\text{in}}}$ into a single reservoir becomes impractical. The number of connections required to maintain expressive internal dynamics scales unfavorably with input dimensionality, leading to increased computational cost and degraded performance due to overloading the reservoir. To address this limitation, \cite{pathak2018model} suggested employing a parallel architecture in which the input space is partitioned into $n$ spatially localized subdomains, each assigned to an independent reservoir. Each reservoir $\mathcal{R}^{(i)}$, with $i\in[1,...n]$, is responsible for modeling a portion of the system's spatial domain and receives as input only a subset of the full signal $\mathbf{u}(t)$, corresponding to its assigned region. To preserve spatial continuity and ensure that the local dynamics are informed by their surroundings, neighboring reservoirs are usually allowed to share overlapping input regions. Specifically, each reservoir's input is extended by a symmetric buffer zone around its central region, allowing it to incorporate information from adjacent spatial components. This input mixing ensures that the model can learn local interactions while maintaining coherence across reservoir boundaries. Since all reservoirs operate independently, each with its own internal dynamics and set of fixed weights, the decomposition enables the application of reservoir computing to large-scale systems by distributing the learning across multiple smaller, localized models. Moreover, it lends itself naturally to parallel implementation, as the reservoirs can be trained and executed independently given the appropriate input structure.

\begin{figure*}[t]
    \centering
    \includegraphics[width=\textwidth]{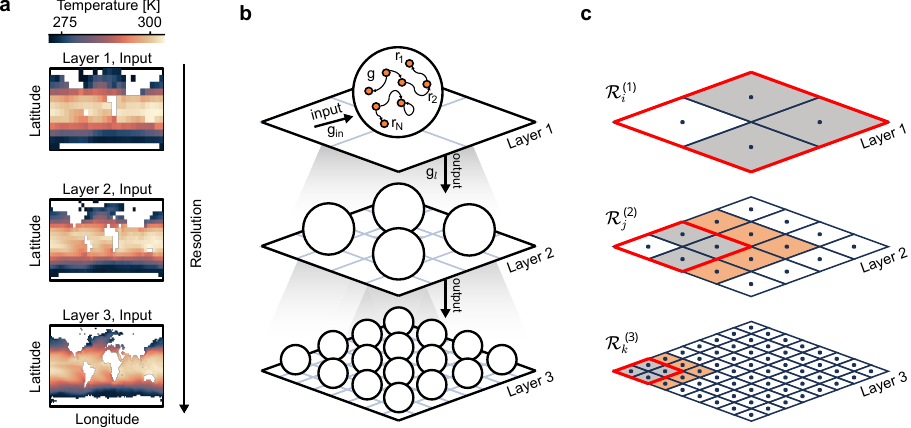}
    \caption{
    \textbf{a,} A dataset is organized into different resolution levels, forming a hierarchical architecture.
    \textbf{b,} Sketch of a three-layer hierarchical model architecture.
    Each resolution level compose a two-dimensional spatial domain represented as a grid of coordinates, each corresponding to an independent time series of a dynamical variable. The domain is partitioned into local subregions each serving as the learning target for a distinct reservoir network.
    Information from distant regions is passed through the previous layer at a lower resolution. In particular, looking at panel \textbf{c}, In each layer, the highlighted reservoir $\mathcal{R}$ (superscript indicates layer hierarchy) receives additional input from the output of the corresponding reservoir in the layer above (red contour), excluding contributions from the region that overlaps directly with its own coordinates (white area). This ensures that only non-local coarse-grained dynamics are passed downward, promoting hierarchical context without redundancy. The orange area highlights the neighboring coordinates which are included as additional input features to incorporate local spatial context as in \cite{pathak2018model}.
    }
    \label{fig:1}
\end{figure*}

\subsection{Cross-Scale Reservoir Computing}
In order to incorporate more efficiently information across very different spatial scales, we propose a novel hierarchical and multilayer approach consisting in representing the same system with a family of grids with different resolution levels from coarser to finer at increasing grid index \(\ell\):  each layer \(\ell\) operates on a regridded version of the input $\mathbf u^{(\ell)} (t)$ with resolution increasing with \(\ell\) (\FigRef{fig:1}a-b).

At a fixed resolution level \(\ell\), each cell of the grid represents a reservoir that evolves and predicts its assigned subset of \(\mathbf{u}^{(\ell)}(t)\), using the previously introduced parallel formulation. Input mixing within each layer is maintained through local spatial overlap, ensuring that each reservoir receives contextual information from its neighboring regions at the same resolution (\FigRef{fig:1}c). 

In addition, hierarchical coupling is introduced by augmenting the input of each reservoir at level \(\ell\) with predictions from the corresponding region in the coarser layer \(\ell-1\). To avoid redundancy and ensure a true multiscale representation, only the nonoverlapping portion of the coarser reservoir output is included, that is, information from the parent region is incorporated excluding the subregion that coincides with the finer layer's own spatial support. As illustrated in \FigRef{fig:1}c, this design ensures that fine-scale reservoirs are informed by broader-scale dynamics without directly duplicating local information already captured at their own resolution. 

The architecture is trained in a top-down manner, beginning with the coarsest layer \(\ell = 1\), and progressing sequentially to finer resolutions. During prediction, the layers are evaluated recursively: at each time step, \(\mathbf{v}^{(\ell)}(t)\) is computed using both the local mixed input from \(\mathbf{u}^{(\ell)}(t)\) and the hierarchical input passed down from \(\mathbf{v}^{(\ell-1)}(t)\). This layered formulation allows to effectively decouple each layer dynamics and deload the computational cost of running a huge number of reservoir nodes in parallel by distributing the learning task across multiple resolutions. Conceptually, the approach parallels classical multiscale methods for gravitational \(N\)-body simulations \cite{barnes1986hierarchical}, which simplify long-range interactions by approximating sufficiently distant particle groups with a single pseudoparticle located at their center of mass.

In our framework, coarse-graining is instead realized through spatial averaging of the input field \(\mathbf{u}(t)\) over increasingly large regridding cells. This process serves as a smoothing operator that attenuates high-frequency spatial variations, isolating the dominant large-scale structures of the system. The coarsest layers therefore act as low-pass filters, capturing slowly varying background dynamics, while finer layers focus on short-range corrections and residual variability. Crucially, by feeding only the nonoverlapping coarse-scale predictions downward, finer reservoirs receive large-scale contextual information without reintroducing low-frequency modes already present in their local input. This coarse-to-fine coupling mechanism allows each reservoir layer to specialize in a distinct spectral band of the system's dynamics, allowing the model to capture interactions across scales. 

\section{Model performances and comparison}

We trained and evaluated our cross-scale reservoir computing model using the Copernicus satellite‐derived global daily sea surface temperature (SST) \cite{copernicus2019sst} spanning from 1981 to 2016.

We fixed the readout nonlinearity $\psi(r) = r$, so that predictive capacity stems linearly from the internal reservoir dynamics. 
Inputs are coarse-grained via spatial averaging into a three-layer resolutions $18^\circ \times 18^\circ$ (layer 1), $6^\circ \times 6^\circ$ (layer 2), and $2^\circ \times 2^\circ$ (layer 3), yielding progressively smoothed fields that emphasize large-scale variability at upper layers and residual fine-scale variability at lower layers.
Each layer is partitioned into overlapping local subregions (tiles, gray area in \FigRef{fig:1}c) served by independent parallel reservoirs. For all the layers the overlap is fixed to 2 tiles.

\begin{figure}[t]
    \centering
    \includegraphics[width=\linewidth]{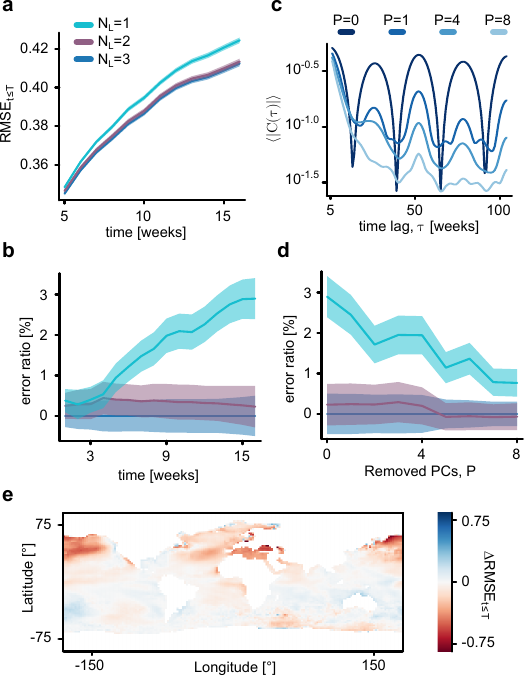}
    \caption{
    \textbf{a,} Average total error (\(\mathrm{RMSE}_{t\le T}\)) as a function of time $T$ for different numbers of layers, \(N_L\).
    \textbf{b,} Ratio of the \(\mathrm{RMSE}_{t\le T}\) for various \(N_L\) to the reference case with \(N_L = 3\), shown as a function of time.
    \textbf{c,} Absolute autocorrelation of the dataset versus time lag, averaged over all spatial locations, for different numbers of removed principal components, \(P\).
    \textbf{d,} Ratio of the \(\mathrm{RMSE}_{t\le T}\) at a fixed time of 10 weeks, plotted as a function of \(P\).
    \textbf{e,} Difference in \(\mathrm{RMSE}_{t \leq T}\) between \(N_L = 3\) and \(N_L = 1\) for individual geographical locations at \(T = 50\) weeks, averaged over 25 runs.  
    In all panels, solid lines denote the mean and shaded areas indicate the SEM, computed over $30$ independent runs (unless otherwise stated).
    }
    \label{fig:2}
\end{figure}

The baseline model consists of a single-layer of parallel reservoirs at fixed resolution, comprising $162$ networks. In contrast, the proposed cross-scale architecture employs a multilayered design with $2$, $18$, and $162$ reservoirs in layers one, two, and three, respectively. Each reservoir contains $1250$ neurons.

We ran a grid search, sweeping g and noise level linearly, and $g_{\mathrm{in}}, g_l,$ and $\tau$ logarithmically across layers.
Better hyperparameter combinations may exist, but the comparison is consistent, exploring the same set of parameters for the highest spatial resolution in the 1-, 2-, and 3-layers models.
For each configuration and prediction horizon \(T\), we computed the total root-mean-square error \(\mathrm{RMSE}_{t \le T}\) defined as
\begin{equation}
\mathrm{RMSE}_{t\le T} = 
\sqrt{
\frac{1}{C}\sum_{i=1}^{C}
\left\langle \big(\mathrm v_i(t) - \hat{\mathrm v}_i(t)\big)^2 \right\rangle_t
} \, ,
\end{equation}
where the temporal average is over all the samples up to time $T$, $C$ is the set of observed coordinates (e.g., grid points), $\hat{\mathrm v}_i(t)$ is the true value, and ${\mathrm v}_i(t)$ is the model forecast for coordinate $i$ at time step $t$.
We defined the best set of parameters the one that correspond with the minimal error for $T=1,5,10,50$ weeks. 

Both two- and three-layer models progressively reduce $\mathrm{RMSE}_{t\le T}$ in time compared to the single-layer baseline, with the gain widening at longer time horizons where error accumulation is most severe \FigRef{fig:2}a.
Error ratios relative to the three-layer model demonstrate that improvements largely saturate between two and three layers, indicating that once dominant slow modes are captured by a layer, further depth does not yield significant returns for this task (\FigRef{fig:2}b).

To pinpoint the source of this improvement, we conducted a principal component analysis (PCA). We removed the first $P$ principal components (PCs) of the target, which carry slowly varying and quasi-periodic variability; the absolute autocorrelation of the data decays faster and with less periodicity as $P$ increases, confirming the removal of some long-memory content (\FigRef{fig:2}c). 
Consistently, the performance gap between multi-layer and single-layer models shrinks as $P$ grows, with the ratio at a fixed horizon of $10$ weeks showing clear degradation of the multi-layer advantage when slow PCs are removed (\FigRef{fig:2}d), demonstrating that inter-layers coupling primarily benefits forecasting by capturing and propagating slow coherent modes.

Finally, to characterize where this multi-layer structure yields the largest forecasting improvements, \FigRef{fig:2}e maps the total error reduction at $50$ weeks when comparing three-layer versus single-layer models $(\Delta \mathrm{RMSE}_{t\le T}=\mathrm{RMSE}_{t\le T}^{N_L{=}3}-\mathrm{RMSE}_{t\le T}^{N_L{=}1})$. Red colors indicate regions where the multi-layer approach performs better, while blue colors show regions where it performs worse. The improvement is most pronounced over mid- and high-latitude oceans of the Northern Hemisphere, while the Southern Hemisphere shows only scattered improvements. This hemispheric difference likely reflects fundamental contrasts in climate dynamics between the two hemispheres. The Northern Hemisphere contains large landmasses that create strong temperature contrasts between land and ocean, promoting hemispheric asymmetry and intensified slowly-evolving climate patterns that persist for months to decades \cite{ruela2020, liu2024} -- exactly the type of data components that benefit from the approach used here.

\section{Emerging linearity} 
\begin{figure}
    \centering
    \includegraphics[width=\linewidth]{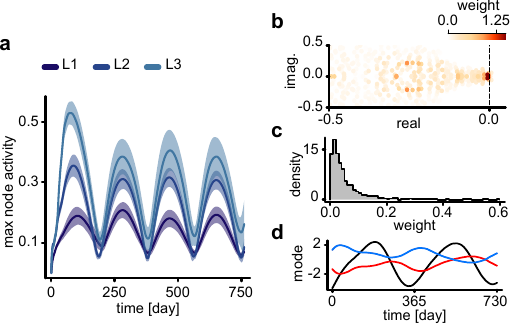}
    \caption{
    \textbf{a,} Maximum network activity across all nodes for three different layers (L1,L2,L3). The solid line represents the mean and the shaded area represents the standard deviation across 25 independent system realizations.
    \textbf{b,} The eigenvalue spectrum of the effective reservoir that reproduces the dynamics of Layer 1 (L1). The color of each point indicates its decomposition weight. The data shown is from a single system realization.
    \textbf{c,} Decomposition weights distribution for the effective reservoir shown in (b).
    \textbf{d,} The real (black) and imaginary (red, blue) components of the complex coniugate linear mode pair associated with the highest decomposition weight.
    }
    \label{fig:3}
\end{figure}

Spatial averaging in the first layers contracts the dynamic range of the reservoir input. Consequently, we expect simpler dynamical regimes for optimally trained networks at lower resolutions. Empirically, we find that the maximum network activity within the optimized reservoirs decreases monotonically from the finest to the coarsest layer (\FigRef{fig:3}a), indicating operation closer to the linear region of the gain function (\EqRef{eq:reservoir-dynamics-2}) at larger spatial scales.
The mechanics of these regimes are fully tractable \cite{di2024linearizing}, enabling the identification of slow dynamical modes that are inherited from the low-resolution input layers.
This linearization yields the approximated autonomous linear dynamics:
\begin{equation}
\tau \dot{\mathbf r} \approx (\mathbf{W} + \mathbf{W}_{\text{in}} \mathbf{W}_{\text{out}}) \mathbf{r} - \mathbf{r} \, .
\label{eq:lrnn_closed}
\end{equation}
Consequently, the dynamical repertoire of the network and its predictive output are determined by the eigenspectrum of the effective connectivity matrix, $\mathbf{\tilde{W}} = \mathbf{W} + \mathbf{W}_{\text{in}} \mathbf{W}_{\text{out}}$.

Diagonalizing $\mathbf{\tilde{W}} = \mathbf{Z} \mathbf{\Lambda} \mathbf{Z}^{-1}$ with $\mathbf{\Lambda}=\mathrm{diag}(\lambda_1,\dots,\lambda_{D_r})$, and introducing coordinates $z(t)=Z^{-1} r(t)$, we obtain independent modal evolutions: $z_k(t) \;=\; z_k(0)\,\exp[(\lambda_k - 1)t / \tau]$. This provides a linear decomposition for the network output signal:
\begin{equation}
\mathrm v_i(t) \approx \sum_{k=1}^{D_r} [\mathbf{W}_{\text{out}} \mathbf{Z}]_{ik} \, z_k(0) e^{(\lambda_k - 1) t / \tau} \, .
\label{eq:linear-dec}
\end{equation}

At the layer level, the ensemble of trained reservoirs can be represented as a single, larger effective reservoir, whose state vector concatenates the states of individual networks and whose structured connectivity matrix reflects interactions induced by overlapping spatial tiles. \FigRef{fig:3}b shows an example of the eigenvalue spectrum of the effective reservoir for the first layer. The eigenvalues are color-coded by their corresponding mode weight ($w_k = \sum _i[\mathrm{W}_{\text{out}} \mathrm{Z}]_{ik}\,  z_k(0)$), which quantifies the contribution of each mode to the output. 
The spectrum exhibits a highly skewed contribution profile: a small subset of modes carries decomposition weights orders of magnitude larger than the bulk, demonstrating that the output is dominated by a few slow oscillatory components (\FigRef{fig:3}c).

The top pair forms a complex conjugate with large weight, generating the leading oscillatory signal that aligns with the annual seasonal cycle in SST, consistent with the slow coherent structure emphasized by coarse-graining (\FigRef{fig:3}d).

\section{Conclusions and Future Directions}

We have introduced a cross-scale reservoir computing framework for large multi-resolution spatiotemporal data. By hierarchically coupling coarse and fine representations, the architecture better captures slow dynamical modes and distributes predictive load across layers. Our results demonstrate consistent improvements over single-layer baselines across forecasting horizons, particularly due to the model capacity to retain and propagate quasi-periodic principal components. The benefits diminish when these slow modes are removed, underscoring their critical role. Additionally, the emergence of linear dynamics in coarser layers allows for interpretable modal decompositions of the learned dynamics.

Beyond accuracy, the hierarchical design confers computational efficiency: by distributing the information across layers, it becomes possible to construct much smaller individual reservoirs compared to large monolithic models that are often computationally intractable. Moreover, tuning and training remain modular and relatively lightweight, as each layer is optimized sequentially in a top-down fashion. Only the highest-resolution layers require substantial tuning effort, which limits computational demands to the finer scales.

Furthermore, the model is easily extensible: it can integrate recently developed, more expressive single-reservoir models, potentially yielding additional performance and interpretability gains \cite{pathak2018hybrid,gauthier2021next, chepuri2024hybridizing, brenner2024almost}.

In this study, we restricted our input to sea surface temperature, omitting potentially informative variables such as atmospheric pressure, wind, or humidity. Extending the analysis to additional datasets could further reveal both strengths and limitations of the approach. 

A promising direction is the integration of spatially adaptive normalization or resolution-sensitive weighting schemes. As recent high-resolution studies show, climatological variability is often highly localized, and predictive architectures should account for both intensity scaling and resolution distance between sensors or grid points \cite{pecorino2024empirical}. Incorporating such spatial sensitivity and, where available physical priors, could further improve performance on irregular and nonstationary domains.

Finally, the cross-scale design is well suited to other domains with naturally hierarchical measurements, including neuroscience. Coupling slow signals (e.g. fMRI) with faster (e.g. MEG, EEG) and fine-grained electrophysiology (e.g. LFP, MUA) \cite{yen2023exploring} may both improve predictive capacity and provide interpretable links between dynamics across spatial scales.

\balance

\bibliography{RefLibrary}

\begin{thebibliography}{24}
\expandafter\ifx\csname natexlab\endcsname\relax\def\natexlab#1{#1}\fi
\providecommand{\url}[1]{\texttt{#1}}
\providecommand{\href}[2]{#2}
\providecommand{\path}[1]{#1}
\providecommand{\DOIprefix}{doi:}
\providecommand{\ArXivprefix}{arXiv:}
\providecommand{\URLprefix}{URL: }
\providecommand{\Pubmedprefix}{pmid:}
\providecommand{\doi}[1]{\href{http://dx.doi.org/#1}{\path{#1}}}
\providecommand{\Pubmed}[1]{\href{pmid:#1}{\path{#1}}}
\providecommand{\bibinfo}[2]{#2}
\ifx\xfnm\relax \def\xfnm[#1]{\unskip,\space#1}\fi
\bibitem[{Arcomano et~al.(2020)Arcomano, Szunyogh, Pathak, Wikner, Hunt and Ott}]{arcomano2020atmospheric}
\bibinfo{author}{Arcomano, T.}, \bibinfo{author}{Szunyogh, I.}, \bibinfo{author}{Pathak, J.}, \bibinfo{author}{Wikner, A.}, \bibinfo{author}{Hunt, B.R.}, \bibinfo{author}{Ott, E.}, \bibinfo{year}{2020}.
\newblock \bibinfo{title}{A machine learning-based global atmospheric forecast model}.
\newblock \bibinfo{journal}{Geophys. Res. Lett.} \bibinfo{volume}{47}, \bibinfo{pages}{e2020GL087776}.
\bibitem[{Barnes and Hut(1986)}]{barnes1986hierarchical}
\bibinfo{author}{Barnes, J.}, \bibinfo{author}{Hut, P.}, \bibinfo{year}{1986}.
\newblock \bibinfo{title}{A hierarchical o (n log n) force-calculation algorithm}.
\newblock \bibinfo{journal}{nature} \bibinfo{volume}{324}, \bibinfo{pages}{446--449}.
\bibitem[{Bauer et~al.(2015)Bauer, Thorpe and Brunet}]{bauer2015quiet}
\bibinfo{author}{Bauer, P.}, \bibinfo{author}{Thorpe, A.}, \bibinfo{author}{Brunet, G.}, \bibinfo{year}{2015}.
\newblock \bibinfo{title}{The quiet revolution of numerical weather prediction}.
\newblock \bibinfo{journal}{Nature} \bibinfo{volume}{525}, \bibinfo{pages}{47--55}.
\bibitem[{Brenner et~al.(2024)Brenner, Hemmer, Monfared and Durstewitz}]{brenner2024almost}
\bibinfo{author}{Brenner, M.}, \bibinfo{author}{Hemmer, C.J.}, \bibinfo{author}{Monfared, Z.}, \bibinfo{author}{Durstewitz, D.}, \bibinfo{year}{2024}.
\newblock \bibinfo{title}{Almost-linear rnns yield highly interpretable symbolic codes in dynamical systems reconstruction}.
\newblock \bibinfo{journal}{Advances in Neural Information Processing Systems} \bibinfo{volume}{37}, \bibinfo{pages}{36829--36868}.
\bibitem[{Chepuri et~al.(2024)Chepuri, Amzalag, Antonsen and Girvan}]{chepuri2024hybridizing}
\bibinfo{author}{Chepuri, R.}, \bibinfo{author}{Amzalag, D.}, \bibinfo{author}{Antonsen, T.}, \bibinfo{author}{Girvan, M.}, \bibinfo{year}{2024}.
\newblock \bibinfo{title}{Hybridizing traditional and next-generation reservoir computing to accurately and efficiently forecast dynamical systems}.
\newblock \bibinfo{journal}{Chaos: An Interdisciplinary Journal of Nonlinear Science} \bibinfo{volume}{34}.
\bibitem[{{Copernicus Climate Change Service, Climate Data Store}(2019)}]{copernicus2019sst}
\bibinfo{author}{{Copernicus Climate Change Service, Climate Data Store}}, \bibinfo{year}{2019}.
\newblock \bibinfo{title}{Sea surface temperature daily gridded data from 1981 to 2016 derived from a multi-product satellite-based ensemble}.
\newblock \bibinfo{howpublished}{\url{https://doi.org/10.24381/cds.ab204534}}.
\newblock \DOIprefix\doi{10.24381/cds.ab204534}. \bibinfo{note}{accessed on 01-09-2024}.
\bibitem[{Di~Antonio et~al.(2024)Di~Antonio, Gili, Gabrielli and Mattia}]{di2024linearizing}
\bibinfo{author}{Di~Antonio, G.}, \bibinfo{author}{Gili, T.}, \bibinfo{author}{Gabrielli, A.}, \bibinfo{author}{Mattia, M.}, \bibinfo{year}{2024}.
\newblock \bibinfo{title}{Linearizing and forecasting: a reservoir computing route to digital twins of the brain}.
\newblock \bibinfo{journal}{bioRxiv} \DOIprefix\doi{10.1101/2024.10.22.619672}.
\bibitem[{Duraisamy et~al.(2019)Duraisamy, Iaccarino and Xiao}]{duraisamy2019turbulence}
\bibinfo{author}{Duraisamy, K.}, \bibinfo{author}{Iaccarino, G.}, \bibinfo{author}{Xiao, H.}, \bibinfo{year}{2019}.
\newblock \bibinfo{title}{Turbulence modeling in the age of data}.
\newblock \bibinfo{journal}{Annual review of fluid mechanics} \bibinfo{volume}{51}, \bibinfo{pages}{357--377}.
\bibitem[{Gallicchio et~al.(2017)Gallicchio, Micheli and Pedrelli}]{gallicchio2017deep}
\bibinfo{author}{Gallicchio, C.}, \bibinfo{author}{Micheli, A.}, \bibinfo{author}{Pedrelli, L.}, \bibinfo{year}{2017}.
\newblock \bibinfo{title}{Deep reservoir computing: A critical experimental analysis}.
\newblock \bibinfo{journal}{Neurocomputing} \bibinfo{volume}{268}, \bibinfo{pages}{87--99}.
\bibitem[{Gauthier et~al.(2021)Gauthier, Bollt, Griffith and Barbosa}]{gauthier2021next}
\bibinfo{author}{Gauthier, D.J.}, \bibinfo{author}{Bollt, E.}, \bibinfo{author}{Griffith, A.}, \bibinfo{author}{Barbosa, W.A.}, \bibinfo{year}{2021}.
\newblock \bibinfo{title}{Next generation reservoir computing}.
\newblock \bibinfo{journal}{Nat. Commun.} \bibinfo{volume}{12}, \bibinfo{pages}{1--8}.
\bibitem[{Jaeger and Haas(2004)}]{Jaeger2004}
\bibinfo{author}{Jaeger, H.}, \bibinfo{author}{Haas, H.}, \bibinfo{year}{2004}.
\newblock \bibinfo{title}{{Harnessing nonlinearity: predicting chaotic systems and saving energy in wireless communication.}}
\newblock \bibinfo{journal}{Science} \bibinfo{volume}{304}, \bibinfo{pages}{78--80}.
\newblock \DOIprefix\doi{10.1126/science.1091277}.
\bibitem[{Kalnay(2003)}]{kalnay2003atmospheric}
\bibinfo{author}{Kalnay, E.}, \bibinfo{year}{2003}.
\newblock \bibinfo{title}{Atmospheric modeling, data assimilation and predictability}.
\newblock \bibinfo{publisher}{Cambridge university press}.
\bibitem[{Kim and Bassett(2023)}]{kim2023neural}
\bibinfo{author}{Kim, J.Z.}, \bibinfo{author}{Bassett, D.S.}, \bibinfo{year}{2023}.
\newblock \bibinfo{title}{A neural machine code and programming framework for the reservoir computer}.
\newblock \bibinfo{journal}{Nature Machine Intelligence} \bibinfo{volume}{5}, \bibinfo{pages}{622--630}.
\bibitem[{Liu et~al.(2024)Liu, Song and Luo}]{liu2024}
\bibinfo{author}{Liu, F.}, \bibinfo{author}{Song, F.}, \bibinfo{author}{Luo, Y.}, \bibinfo{year}{2024}.
\newblock \bibinfo{title}{Human-induced intensified seasonal cycle of sea surface temperature}.
\newblock \bibinfo{journal}{Nature Communications} \bibinfo{volume}{15}, \bibinfo{pages}{3948}.
\bibitem[{Luko{\v{s}}evi{\v{c}}ius and Jaeger(2009)}]{lukovsevivcius2009reservoir}
\bibinfo{author}{Luko{\v{s}}evi{\v{c}}ius, M.}, \bibinfo{author}{Jaeger, H.}, \bibinfo{year}{2009}.
\newblock \bibinfo{title}{Reservoir computing approaches to recurrent neural network training}.
\newblock \bibinfo{journal}{Comput. Sci. Rev.} \bibinfo{volume}{3}, \bibinfo{pages}{127--149}.
\bibitem[{Manneschi et~al.(2021)Manneschi, Ellis, Gigante, Lin, Del~Giudice and Vasilaki}]{manneschi2021exploiting}
\bibinfo{author}{Manneschi, L.}, \bibinfo{author}{Ellis, M.O.}, \bibinfo{author}{Gigante, G.}, \bibinfo{author}{Lin, A.C.}, \bibinfo{author}{Del~Giudice, P.}, \bibinfo{author}{Vasilaki, E.}, \bibinfo{year}{2021}.
\newblock \bibinfo{title}{Exploiting multiple timescales in hierarchical echo state networks}.
\newblock \bibinfo{journal}{Frontiers in Applied Mathematics and Statistics} \bibinfo{volume}{6}, \bibinfo{pages}{616658}.
\bibitem[{Pathak et~al.(2018a)Pathak, Hunt, Girvan, Lu and Ott}]{pathak2018model}
\bibinfo{author}{Pathak, J.}, \bibinfo{author}{Hunt, B.}, \bibinfo{author}{Girvan, M.}, \bibinfo{author}{Lu, Z.}, \bibinfo{author}{Ott, E.}, \bibinfo{year}{2018}a.
\newblock \bibinfo{title}{Model-free prediction of large spatiotemporally chaotic systems from data: A reservoir computing approach}.
\newblock \bibinfo{journal}{Phys. Rev. Lett.} \bibinfo{volume}{120}, \bibinfo{pages}{024102}.
\bibitem[{Pathak et~al.(2018b)Pathak, Wikner, Fussell, Chandra, Hunt, Girvan and Ott}]{pathak2018hybrid}
\bibinfo{author}{Pathak, J.}, \bibinfo{author}{Wikner, A.}, \bibinfo{author}{Fussell, R.}, \bibinfo{author}{Chandra, S.}, \bibinfo{author}{Hunt, B.R.}, \bibinfo{author}{Girvan, M.}, \bibinfo{author}{Ott, E.}, \bibinfo{year}{2018}b.
\newblock \bibinfo{title}{Hybrid forecasting of chaotic processes: Using machine learning in conjunction with a knowledge-based model}.
\newblock \bibinfo{journal}{Chaos: An interdisciplinary journal of nonlinear science} \bibinfo{volume}{28}.
\bibitem[{Pecorino et~al.(2024)Pecorino, Matteo, Milazzo, Pasotti, Pluchino and Rapisarda}]{pecorino2024empirical}
\bibinfo{author}{Pecorino, V.}, \bibinfo{author}{Matteo, T.D.}, \bibinfo{author}{Milazzo, M.}, \bibinfo{author}{Pasotti, L.}, \bibinfo{author}{Pluchino, A.}, \bibinfo{author}{Rapisarda, A.}, \bibinfo{year}{2024}.
\newblock \bibinfo{title}{Empirical analysis of hourly rainfall data in sicily from 2002 to 2023}.
\newblock \bibinfo{journal}{The European Physical Journal B} \bibinfo{volume}{97}, \bibinfo{pages}{154}.
\bibitem[{Price et~al.(2025)Price, Sanchez-Gonzalez, Alet, Andersson, El-Kadi, Masters, Ewalds, Stott, Mohamed, Battaglia et~al.}]{price2025probabilistic}
\bibinfo{author}{Price, I.}, \bibinfo{author}{Sanchez-Gonzalez, A.}, \bibinfo{author}{Alet, F.}, \bibinfo{author}{Andersson, T.R.}, \bibinfo{author}{El-Kadi, A.}, \bibinfo{author}{Masters, D.}, \bibinfo{author}{Ewalds, T.}, \bibinfo{author}{Stott, J.}, \bibinfo{author}{Mohamed, S.}, \bibinfo{author}{Battaglia, P.}, et~al., \bibinfo{year}{2025}.
\newblock \bibinfo{title}{Probabilistic weather forecasting with machine learning}.
\newblock \bibinfo{journal}{Nature} \bibinfo{volume}{637}, \bibinfo{pages}{84--90}.
\bibitem[{Raissi et~al.(2019)Raissi, Perdikaris and Karniadakis}]{raissi2019physics}
\bibinfo{author}{Raissi, M.}, \bibinfo{author}{Perdikaris, P.}, \bibinfo{author}{Karniadakis, G.E.}, \bibinfo{year}{2019}.
\newblock \bibinfo{title}{Physics-informed neural networks: A deep learning framework for solving forward and inverse problems involving nonlinear partial differential equations}.
\newblock \bibinfo{journal}{Journal of Computational physics} \bibinfo{volume}{378}, \bibinfo{pages}{686--707}.
\bibitem[{Ruela et~al.(2020)Ruela, Sousa, DeCastro and Dias}]{ruela2020}
\bibinfo{author}{Ruela, R.}, \bibinfo{author}{Sousa, M.}, \bibinfo{author}{DeCastro, M.}, \bibinfo{author}{Dias, J.}, \bibinfo{year}{2020}.
\newblock \bibinfo{title}{Global and regional evolution of sea surface temperature under climate change}.
\newblock \bibinfo{journal}{Global and Planetary Change} \bibinfo{volume}{190}, \bibinfo{pages}{103190}.
\bibitem[{Yan et~al.(2024)Yan, Huang, Bienstman, Tino, Lin and Sun}]{yan2024emerging}
\bibinfo{author}{Yan, M.}, \bibinfo{author}{Huang, C.}, \bibinfo{author}{Bienstman, P.}, \bibinfo{author}{Tino, P.}, \bibinfo{author}{Lin, W.}, \bibinfo{author}{Sun, J.}, \bibinfo{year}{2024}.
\newblock \bibinfo{title}{Emerging opportunities and challenges for the future of reservoir computing}.
\newblock \bibinfo{journal}{Nature Communications} \bibinfo{volume}{15}, \bibinfo{pages}{2056}.
\bibitem[{Yen et~al.(2023)Yen, Lin and Chiang}]{yen2023exploring}
\bibinfo{author}{Yen, C.}, \bibinfo{author}{Lin, C.L.}, \bibinfo{author}{Chiang, M.C.}, \bibinfo{year}{2023}.
\newblock \bibinfo{title}{Exploring the frontiers of neuroimaging: a review of recent advances in understanding brain functioning and disorders}.
\newblock \bibinfo{journal}{Life} \bibinfo{volume}{13}, \bibinfo{pages}{1472}.

\end{thebibliography}

\end{document}